\documentclass[journal]{IEEEtran}
\usepackage{cite}

\ifCLASSINFOpdf
\else
\fi
\usepackage{amsmath}
\usepackage{amssymb}

\usepackage{url}

\usepackage{subcaption}
\usepackage{graphicx}
\usepackage{xcolor}
\usepackage{multirow,booktabs}
\usepackage{listings}
\usepackage{multicol} 

\definecolor{codegreen}{rgb}{0,0.6,0}
\definecolor{codegray}{rgb}{0.5,0.5,0.5}
\definecolor{codepurple}{rgb}{0.58,0,0.82}
\definecolor{backcolour}{rgb}{0.95,0.95,0.92}

\lstdefinestyle{mystyle}{
    backgroundcolor=\color{backcolour},   
    commentstyle=\color{codegreen},
    keywordstyle=\ttfamily,
    numberstyle=\tiny\color{codegray},
    stringstyle=\ttfamily,
    basicstyle=\ttfamily\footnotesize,
    breakatwhitespace=false,         
    breaklines=true,                 
    captionpos=b,                    
    keepspaces=true,                 
    numbers=none,                    
    numbersep=5pt,                  
    showspaces=false,                
    showstringspaces=false,
    showtabs=false,                  
    tabsize=2
}

\begin{document}
%
\title{Efficient Prompting for LLM-based Generative Internet of Things}
%
%
%

\author{Bin Xiao,
        Burak Kantarci,~\IEEEmembership{Senior~Member,~IEEE},  Jiawen Kang,~\IEEEmembership{Senior~Member,~IEEE}\\  Dusit Niyato,~\IEEEmembership{Fellow,~IEEE},  Mohsen Guizani,~\IEEEmembership{Fellow,~IEEE}
        
        \thanks{B. Xiao and B. Kantarci are with the School of Electrical Engineering and Computer Science, University of Ottawa, Ottawa, ON, Canada. Emails:\{bxiao103, burak.kantarci\}@uottawa.ca\\
        J. Kang is with Guandong University of Technology, China, Email: kavinkang@gdut.edu.cn\\
        D. Niyato is with Nanyang Technological University, Singapore, Email: dniyato@ntu.edu.sg\\
        M. Guizani is with Mohamed bin Zayed University of Artificial Intelligence (MBZUAI), Abu Dhabi, UAE, mohsen.guizani@mbzuai.ac.ae
}

        }

%
%

\markboth{}%
{Shell \MakeLowercase{\textit{et al.}}: Bare Demo of IEEEtran.cls for IEEE Journals}
%



\maketitle

\begin{abstract}
Large language models (LLMs) have demonstrated remarkable capacities on various tasks, and integrating the capacities of LLMs into the Internet of Things (IoT) applications has drawn much research attention recently. Due to security concerns, many institutions avoid accessing state-of-the-art commercial LLM services, requiring the deployment and utilization of open-source LLMs in a local network setting. However, open-source LLMs usually have more limitations regarding their performance, such as their arithmetic calculation and reasoning capacities, and practical systems of applying LLMs to IoT have yet to be well-explored. Therefore, we propose an LLM-based Generative IoT (GIoT) system deployed in the local network setting in this study. To alleviate the limitations of LLMs and provide service with competitive performance, we apply prompt engineering methods to enhance the capacities of the open-source LLMs, design a Prompt Management Module and a Post-processing Module to manage the tailored prompts for different tasks and process the results generated by the LLMs. To demonstrate the effectiveness of the proposed system, we discuss a challenging Table Question Answering (Table-QA) task as a case study of the proposed system, as tabular data is usually more challenging than plain text because of their complex structures, heterogeneous data types and sometimes huge sizes. We conduct comprehensive experiments on two popular Table-QA datasets, and the results show that our proposal can achieve competitive performance compared with state-of-the-art LLMs, demonstrating that the proposed LLM-based GIoT system can provide competitive performance with tailored prompting methods and is easily extensible to new tasks without training.
\end{abstract}


%
\IEEEpeerreviewmaketitle

\section{Introduction}
\label{sec:introduction}
%
%
%
%

\IEEEPARstart{A}{rtificial} Intelligence of Things (AIoT), integrating Artificial Intelligence (AI) and Internet of Things (IoT) for efficient data analysis and intelligent decision making~\cite{hu2023adaptive}, have been widely discussed in many studies and applied in many scenarios, such as Healthcare~\cite{adil2024healthcare}, Smart Cities~\cite{fan2023understanding} and Industries~\cite{franco2021survey}. Currently, task-specific machine learning and Deep Neural Network (DNN) models are the mainstream choices for AIoT providing services to IoT devices, which are often deployed on the cloud and edge servers~\cite{hu2023adaptive}. With the development of Generative Artificial Intelligence (GAI), especially large language models (LLMs), leveraging its remarkable general capacities to the IoT applications, termed as Generative Internet of Things (GIoT)~\cite{wen2024generative}, becomes a promising research direction~\cite{wang2024iot, zhong2024casit}. Different from task-specific models, LLMs, such as GPT-4~\cite{OpenAI2023GPT4TR}, have demonstrated their general capacities on a wide range of tasks, such as data analysis, code generation, reasoning, planning and many others, making it possible to provide various services with a single LLM model, maintaining competitive performance with tailored task-specific models.

Despite the remarkable capacities of LLMs, many issues need to be considered when integrating LLMs into IoT systems for an LLM-based GIoT system. First, following the Scaling Law~\cite{kaplan2020scaling}, the capacities of an LLM are growing with its number of parameters and the scale of the training dataset, making LLMs often have a tremendous number of parameters, which leads to high hardware requirements for the training and inference. For example, popular open-source LLMs providing competitive performance on some public benchmarks often have around 70 billion parameters, such as Mixtral-8x7B~\cite{jiang2024mixtral} and Llama-3-70B~\cite{meta2024llama3}. These numbers of parameters make it not practical to deploy LLMs on edge IoT devices. Even though deploying LLMs on the edge and cloud servers can be an option, the efficiency of an LLM-based GIoT system still needs to be carefully considered. Besides the hardware requirements and efficiency issues, data privacy and security issues hinder many institutions from accessing commercial state-of-the-art LLMs~\cite{ray2023chatgpt, yan2024protecting}. For example, for a healthcare IoT application collecting private data from patients, it is necessary to avoid uploading collected data to public, commercial LLM services for data analysis because of privacy issues. Therefore, a safe, transparent, and controllable open-source LLM deployed in a local network is more suitable for many institutions, even though commercial LLMs can provide better services. At last, the performance and the scalability of an LLM-based GIoT system are another two critical considerations because both commercial and open-source LLMs have some inherent limitations~\cite{gao2023pal, chen2023program, wei2022chain}, such as their hallucination issues, limited reasoning capacities for complex tasks. Typically, prompting methods and fine-tuning are two directions that can alleviate these limitations. Specifically, prompting methods improve the performance of an LLM by designing tailored prompts for different tasks to elicit the capacities of an LLM. For example, Chain of Thought (CoT)~\cite{wei2022chain} is a popular prompting method to improve the LLM's reasoning performance by providing reasoning rationales. Program of Thoughts (PoT)~\cite{chen2023program} is another prompting method generating Python code to offload the reasoning and calculation tasks to the Python interpreter. Since these prompting methods focus on tailoring task-specific prompts for LLMs, they often lead to longer inference time because of their larger number of prompting tokens. By contrast, fine-tuning an LLM for a task can also improve the performance but requires labeled datasets and computation resources for the model training. As discussed in some studies~\cite{tu2024towards, yin2023modulora}, fine-tuning an LLM for unseen tasks can increase the model's bias and degrade its general capacities. 

\begin{figure}[htp]
\begin{center}
  \includegraphics[width=\columnwidth]{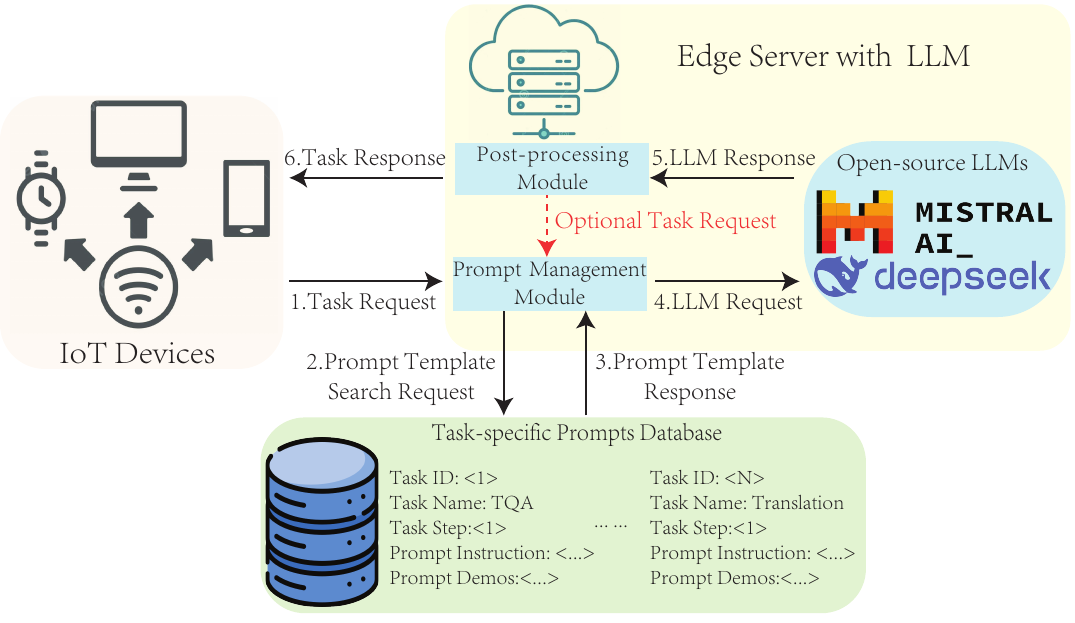}
  \caption{Overall architecture of the proposed LLM-based GIoT system.}
  \label{fig:giot_system}
\end{center}
\end{figure}

Since practical solutions applying LLMs to the IoT setting have yet to be well-explored, even though there have been some studies~\cite{wang2024iot, zhong2024casit} discussing the potential and possible frameworks of such applications, we propose a practical LLM-based GIoT system in this study. Considering the discussed possible issues for an LLM-based GIoT system, we propose to deploy the open-source LLM on the edge server in a local network setting to address the and employ prompting methods to enhance the capacities of the proposed system for different tasks. Specifically, a Prompt Management Module and a Post-processing Module are proposed to be deployed in the edge server, in which the former is responsible for the selection, management and creation of prompts and demonstrations for the requests from IoT devices, and the latter is responsible for post-processing the results generated by the LLMs, as shown in Figure~\ref{fig:giot_system}. With the proposed Prompt Management Module and Post-processing Module, the GIoT system can be easily extended to new tasks by adding tailored task-specific prompts in the Task-specific Prompts Database, providing competitive performance for a wide range of tasks.

To demonstrate the effectiveness of the proposed LLM-based GIoT system, we implement a Semi-structured Table Question Answering (Table-QA) service in the proposed system, which is useful and challenging. Specifically, the Table-QA service aims to answer the query question based on the given table information. Since tables are widely used to summarize critical information in many data sources, such as visually rich documents and web pages~\cite{xiao2023table}, Table-QA~\cite{herzig2020tapas, jiang2022omnitab,zhu2024tat,yu2023towards,zhao2023large} can be a useful service to provide analysis to the tabular data, and has also drawn much research attention recently. Typically, tables can be easily categorized into two groups: structured and semi-structured tables. Structured tables are usually from relational database systems with explicit schema describing their structures and data types, meaning the programming languages, such as SQL, can naturally process them. By contrast, tables from other sources, such as web pages and visually rich documents, are usually semi-structured without schema requirements, resulting in complex structures, heterogeneous data types and sometimes huge sizes, making the Table-QA task on these semi-structured tables more challenging. Therefore, we use the Table-QA problem on the semi-structured tables to verify the proposed LLM-based system, which is a more challenging setting. To provide competitive service as commercial LLMs, we propose a three-stage prompting method, including task-planning, task-conducting and task-correction stages, leveraging Python code to conduct reasoning steps. The proposed prompting method can improve the performance of open-source LLMs, which can also demonstrate that the proposed system can be easily extended to new tasks with prompting methods.


To sum up, the contributions of this study can be 3-fold:

\begin{enumerate}

\item We propose an LLM-based GIoT system with open-source LLMs deployed in the local network setting, which includes a Prompt Management Module, a Post-processing Module and a Task-specific Prompts Database to address the considerations in data privacy and security, system scalability, and enhance the capacities of the LLM by integrating prompting methods. 

\item We discuss a challenging Table-QA problem to demonstrate the proposed LLM-based GIoT system and propose a three-stage prompting solution, including task-planning, task-conducting and task-correction stages, to alleviate the issues caused by the complex structures, heterogeneous data types, huge tables and the limitations of LLMs and also reduce the inference cost. A series of atomic operations is proposed to describe and measure the similarities of QA tasks and select proper demonstrations for prompt creation.

\item We conduct extensive experiments on the WikiTableQA and TabFact datasets with open-source LLMs to verify the proposed LLM-based GIoT system and the proposed prompting method. The experimental results demonstrate that our proposed prompting method can outperform the baseline methods by a large margin, achieving state-of-the-art performance. We also conduct comprehensive analyses that consider the performance of different prompting methods, the inference costs, and the behaviour of different open-source LLMs, which can be a guide for selecting LLMs and prompting methods for an LLM-based GIoT system.

\end{enumerate}
The rest of this paper is organized as follows: Section~\ref{sec:related_work} discusses related studies, including recent studies applying LLMs to the IoT systems, Table-QA solutions and prompting methods. Section~\ref{sec:proposed_method} describes our proposed LLM-based GIoT system and prompting solution. Section~\ref{sec:experiments} shows the experimental results and discusses the design aspects of the proposed prompting method. At last, we draw our conclusion and possible future directions in Section~\ref{sec:conclusion}.

\section{Related Work}
\label{sec:related_work}
\subsection{Generative Models in IoT}
\label{sec:generative_models_in_iot}
As generative models, especially text-based LLMs, have demonstrated their remarkable capacities in a wide range of tasks, integrating their capacities into IoT applications has attracted the research community's attention. There have been some studies~\cite{wen2024generative, wang2024iot, de2022deep} summarizing the critical components of Generative Models and discussing the potential of applying Generative Models to IoT systems. Authors in study~\cite{wen2024generative} firstly term the combination of generative models and IoT applications as Generative IoT (GIoT), discuss the foundations of generative models and the potential of GIoT applications, including Vision-based, Audio-based, Text-based and other GIoT applications. Similarly, authors in study~\cite{wang2024iot} also point out various potential GIoT applications in many fields, such as Mobile Networks, Autonomous Vehicles, and many others. Besides discussing the insights and potentials of GIoT, some studies examine various aspects of applying LLMs in IoT settings. CASIT~\cite{zhong2024casit} is an LLM-agent-based IoT framework proposing a Sensor Interface to convert sensor data into natural language and design multiple types of LLM-based Agent to cooperate and analyze the sensor data and user requests. LLMind~\cite{cui2023llmind} introduces a framework that integrates various domain-specific AI modules and enables IoT device cooperation to enhance the LLM's capabilities for conducting complex tasks. Study~\cite{rong2024leveraging} proposes an intelligent control framework integrating Integrated Terrestrial Non-terrestrial Networks, IoT and language models, identifying key components, potential applications and challenges. Overall, current studies regarding GIoT applications mainly focus on proposing abstractions of the systems, and practical cases need to be explored in more depth. Therefore, this study proposes an extensible LLM-based GIoT system using prompting methods and uses a challenging Table-QA problem as a case study to illustrate the proposed system.

\subsection{Table Question Answering}
\label{sec:related_work_table_qa}
Since this study uses the Table-QA task as the case study of the proposed LLM-based GIoT system, we include recent studies using LLM to solve the Table-QA problem in this section. Specifically, most of the studies applying LLMs to the Table-QA task can be categorized into instruction tuning based and prompt engineering based approaches. Instruction tuning approaches usually need to collect large-scale datasets and then further fine-tuned LLMs with parameter-efficient fine-tuning methods, such as LORA~\cite{hu2021lora} and LongLORA~\cite{chen2023longlora}. TableLLAMA~\cite{zhang2023tablellama} and TAT-LLM~\cite{zhu2024tat} are typical examples of applying instruction tuning for the table processing. TableLLAMA is a generalist model for tables fine-tuned on Llama2~\cite{touvron2023llama2} with LongLORA. For fine-tuning TableLLAMA, a dataset collection named TableInstruct is proposed by collecting table-based samples from 14 datasets for 11 tasks. TAT-LLM is another fine-tuned LLAMA2 model specifically for the discrete reasoning over tables, which decomposes the Table-QA into three steps: Extractor, Reasoner and Executor. To construct the dataset for the model fine-tuning, TAT-LLM proposes a template to guide LLM in generating data following the proposed step-wise pipeline. 

On the other hand, prompting engineering based solutions focus on proposing proper prompts to the language model to elicit LLMs' capacities. Since Table-QA needs to extract relevant information and evidence from tables and perform reasoning, decomposing the Table-QA task into multiple steps is also widely adopted in prompting engineering-based solutions. Besides, as LLM often fails to process large tables and complex reasoning, many studies~\cite{chen2023large, ye2023large, lin2023inner, zhu2024tat} propose to decompose large tables into small tables in the extraction step and divide the complex reasoning task into more straightforward reasoning questions. For example, StructGPT~\cite{Jiang2023StructGPTAG} defines specialized interfaces for information extraction and linearizes the extracted sub-tables as inputs to the LLM for question answering. EEDP~\cite{srivastava2024assessing} is another example proposing a prompting method containing four steps: Elicit, Extract, Decompose and Predict.

Besides these studies decomposing the Table-QA into extraction and reasoning steps and dividing difficult extraction and reasoning tasks into simpler ones, programming languages, such as SQL and Python, are also widely leveraged in these solutions to overcome the limition of LLMs in arithmetic calculation and reasoning. Dater~\cite{ye2023large} is a typical solution following the multi-step design and leveraging SQL to conduct reasoning. More specifically, Dater decomposes both large evidence and complex questions into relevant and simpler ones, applies SQL queries to produce numerical and logical reasoning results to the decomposed sub-questions, and generates the final results based on the sub-results and extracted evidence with ICL. Binding~\cite{cheng2022binding} is another example of applying Python and SQL for the table processing, which proposes a unified API to map tasks into executable programs.

All in all, as the complexities of Table-QA in the information extraction and reasoning, breaking Table-QA into multiple steps and decomposing difficult extraction and reasoning tasks into simpler ones have been the dominant solution, and external tools such as SQL and Python, can compensate the limitations of LLMs in arithmetic calculation and reasoning.

\subsection{Prompting Methods}
This section focuses on recent prompting methods for LLMs because our proposed LLM-based GIoT system applies prompting methods to enhance the LLMs' capacities for different tasks. As some of these methods have also been applied to the Table-QA problem, some methods included in this section can be overlapped with the discussed studies in Section~\ref{sec:related_work_table_qa}. There have been many studies demonstrated that prompting LLMs with a few demonstrations and intermediate reasoning steps can elicit the reasoning capacities of LLMs, which often refer to In Context Learning (ICL)~\cite{brown2020language} and CoT~\cite{wei2022chain}. Although ICL and CoT are useful, writing proper prompting demonstrations is non-trivial and time-consuming. Therefore, some studies~\cite{zhang2023automatic,shao2023synthetic,shum-etal-2023-automatic} propose to use LLMs to generate demonstrations. Auto-CoT~\cite{zhang2023automatic} proposes to prompt LLMs with Zero-shot CoT to generate reasoning rationales but finds that the generated rationales often contain mistakes. To mitigate the wrong demonstration issue, Auto-CoT proposes clustering and sampling the questions first and then applying simple heuristics to sample simpler questions and rationales to construct demonstrations. Synthetic Prompting~\cite{shao2023synthetic} is another typical solution for constructing demonstrations with LLMs, containing forward and backward processes. In the backward process of Synthetic Prompting, a topic word, a target complexity and a self-generated reasoning chain are used as conditions to generate the synthetic question, and the synthetic question is used in the forward process to generate the precise synthetic reasoning chain. Along with these studies focusing on demonstration generation with LLMs, ensemble methods are also effective for reasoning. For example, self-consistency decoding~\cite{wang2022self} first generates a set of candidate outputs from the LLM with different sampling methods, such as temperature sampling, and then aggregates the sampled outputs and uses the most consistent output as the final result. Multi-Chain Reasoning~\cite{yoran-etal-2023-answering} is another ensemble method mixing information from multiple reasoning chains. Different from studies~\cite{wang2022self} to ensemble results of multi-reasoning chains, Multi-Chain Reasoning collects evidence from multiple reasoning chains and prompts the LLM to give the final answer.

In addition to these prompting methods, the integration of programming languages and other external tools holds great potential for overcoming the limitations of LLMs. For instance, LLMs often struggle with precise arithmetic calculations, especially division operations, and staying updated with the latest information. To address these issues, PoT~\cite{chen2023program}, PAL~\cite{gao2023pal}, and many other studies~\cite{cao2023api} leverage Programming Language to assist the reasoning steps. Some studies~\cite{NEURIPS2023_d842425e} introduce the concept of Agent and call external tools by parsing the LLM outputs to enhance the LLM's capabilities.

\section{System Model}
\label{sec:system_model}

\begin{figure*}[htp]
\begin{center}
  \includegraphics[width=\textwidth]{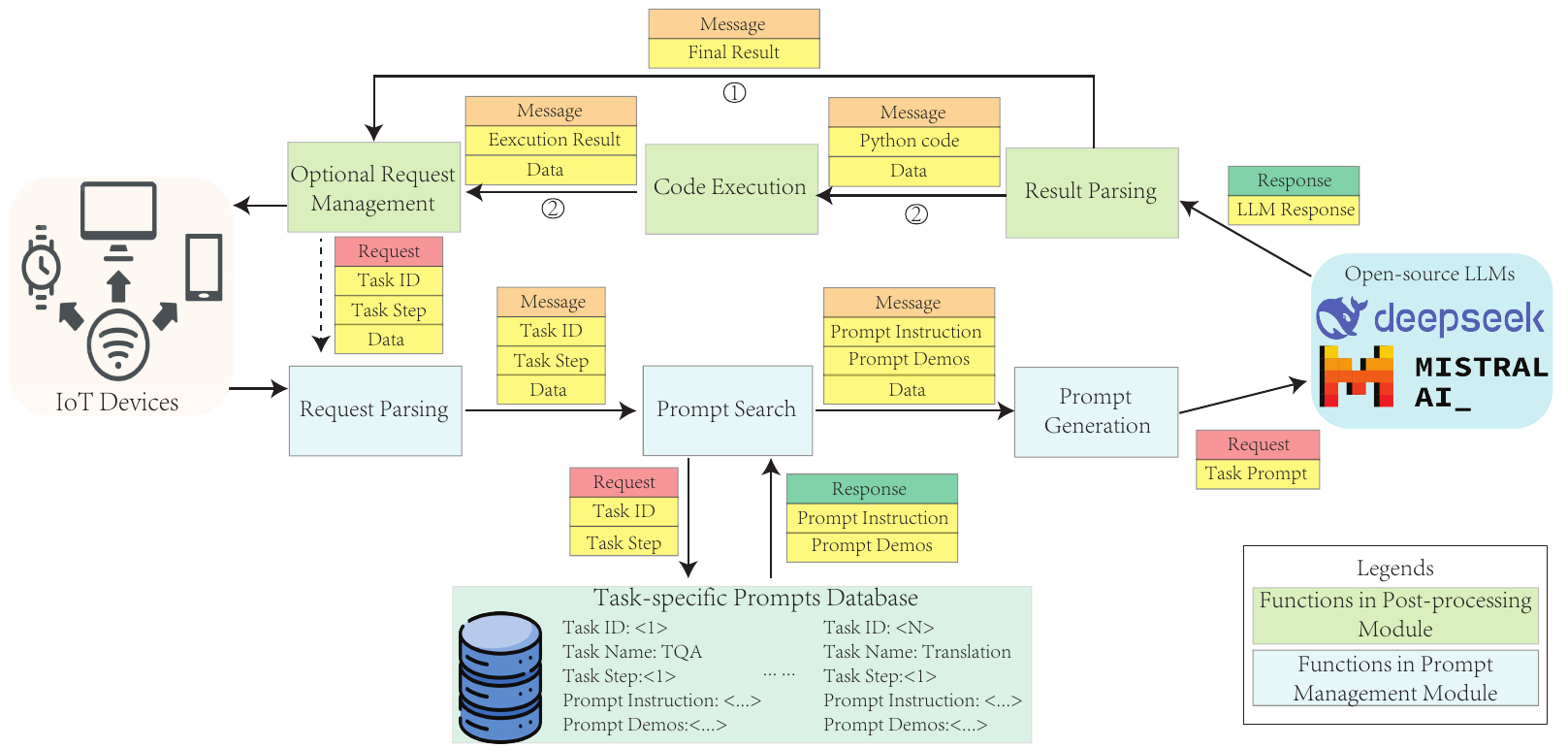}
  \caption{Detailed workflow of the proposed LLM-based GIoT system.}
  \label{fig:detailed_workflow_giot}
\end{center}
\end{figure*}

As discussed in Section~\ref{sec:introduction}, in this study, we consider the scenarios in which IoT devices cannot access commercial LLMs because of data privacy and security considerations, which is a practical setting in many institutions, such as hospitals. Alternatively, as shown in Figure~\ref{fig:giot_system}, an open-source LLM can be deployed in a local edge server to process the requests from IoT devices. Because fine-tuning an LLM for a specific task requires substantial datasets and computational resources, limiting the scalability of extending to multiple tasks, limiting the scalability of extending to multiple tasks, we propose to use prompting methods to enhance the capacities for different tasks with a Prompt Management Module, a Post-processing Module and a Task-specific Prompts Database, as shown in Figure~\ref{fig:giot_system}. Typically, a prompt often consists of instructions and demonstrations, in which the instructions are used to describe the task, and the demonstrations are the examples provided to the LLMs to show the desired outputs. For a text-based request from IoT devices, such as a translation task to a sentence, the proposed Prompt Management Module is responsible for parsing the request and searching for the task-specific prompt instructions and demonstrations regarding the task, as steps 1, 2 and 3 in Figure~\ref{fig:giot_system}. After obtaining the task-specific prompt instructions and demonstrations, the Prompt Management Module constructs the final prompt and sends the request to the LLM. In our design, the capacities to deal with new tasks can be easily extended to the system by adding new prompt instructions and demonstrations in the Task-specific Prompt Database without training or fine-tuning to the LLM~\cite{vatsal2024survey}. Since the results generated by the LLM are sometimes not ideal, we propose a Post-processing Module to process the results further. For example, some methods must prompt the LLM model multiple times to obtain the final results, whose intermediate results and Optional Task Request should be processed by the Post-processing Module. Besides, programming language-aided prompting methods, such as PAL~\cite{gao2023pal} and PoT~\cite{chen2023program}, generate the Python code instead of the final results, which means that the Post-processing Module should also be responsible for executing the generated Python code to obtain the final results. It is worth mentioning that, in the proposed solution, the IoT devices and the edge server can be easily connected with Wi-Fi and other wireless network protocols, and the communication between them can be easily implemented with HTTP protocol.

Following these discussed steps, Figure~\ref{fig:detailed_workflow_giot} shows the detailed components of the proposed two modules and the detailed workflow of the proposed GIoT system. Specifically, the Prompt Management Module consists of three components: Request Parsing, Prompt Search, and Prompt Generation. The Request Parsing component receives and parses the requests from the IoT devices and outputs Task ID, Task Step and parsed Data, in which Task ID and Task Step would be used in the Prompt Search component to search corresponding Prompt Instructions and Prompt Demonstrations from the Task-specific Prompts Database. The parsed Data generated by the Request Parsing component is the information that needs to be further processed by the LLM. For example, for a service of translating English to Chinese, the request from the IoT devices can be \textit{"Task Name: Translation. Data: This is a sample translation service."} Then the Data parsed by the Request Parsing component should be \textit{"This is a sample translation service."}, which would be used by the Prompt Generation component to generate the final Task Prompt together with the Prompt Instruction and Prompt Demonstrations. It is worth mentioning that Prompt Generation can implement customize demonstration selection methods to optimize the performance of applying In Context Learning (ICL)~\cite{brown2020language}. For the Post-processing Module, since the output of the LLM cannot always follow the instructions, a Result Parsing function is needed to refine the outputs. For single-stage prompting methods, the result generated by the Result Parsing function can be the final result returned to the IoT device, as the path 1 shown in Figure~\ref{fig:detailed_workflow_giot}. By contrast, prompting methods, such as PoT~\cite{chen2023program} and PAL~\cite{gao2023pal}, generate Python code and conduct the reasoning steps by an external Python Interpreter. Therefore, a Code Execution component should be included in the Post-processing Module. At last, an Optional Request Management component need to be implemented for the prompting methods with multiple-stages.

\section{Case Study of Table-QA for the LLM-based GIoT System}
\label{sec:proposed_method}
In this section, we use the Table-QA task as a case study to demonstrate the proposed LLM-based GIoT System. We first provide the problem formulation and examine the challenging aspects of the Table-QA task in Section~\ref{sec:problem_analysis}. Then, we describe our proposed three-stage prompting method in detail in Section~\ref{sec:proposed_prompting_method_for_tqa}.
\subsection{Problem Formulation and Analysis}
\label{sec:problem_analysis}
\begin{figure*}[htp]
\begin{center}
  \includegraphics[width=\textwidth]{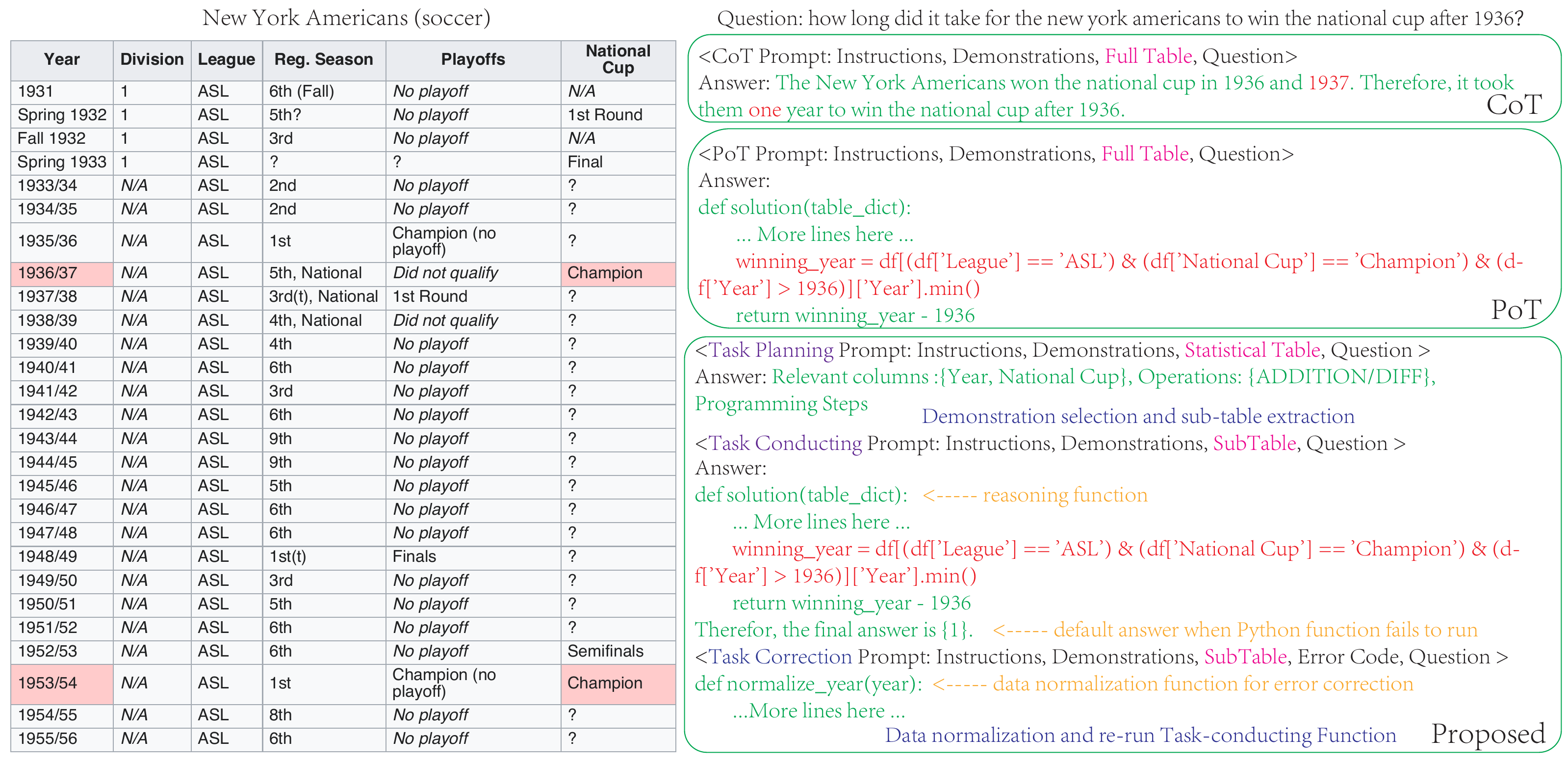}
  \caption{Comparison of CoT, PoT and the proposed method. It is worth mentioning that many details are omitted due to space limitations. The proposed method contains task-planning, task-conducting, and task-correction stages, and it uses a statistical table and sub-tables in these stages to avoid the original large tables.}
  \label{fig:method_comparision_with_benchmarks}
\end{center}
\end{figure*}
A Table-QA service deployed in the proposed LLM-based GIoT system contains two key components: a parametric LLM with parameters $\theta$ and a Retriever $\mathcal{R}$, in which the Retriever $\mathcal{R}$ is responsible for selecting proper instructions and demonstrations used in ICL. Therefore, for an input query pair, $(t, q)$, including a query table and a query question, the output answer after $n$ rounds of prompting $y_n$ can be defined as:
\[y_{i+1} = f_{\theta}(y_i, t, q, \mathcal{R}_i), \quad \text{for } i = 0, 1, \dots, n-1\] where $y_0$ is an empty string. 

\begin{figure}
     \centering
     \begin{subfigure}[b]{\columnwidth}
         \centering
         \includegraphics[width=\columnwidth]{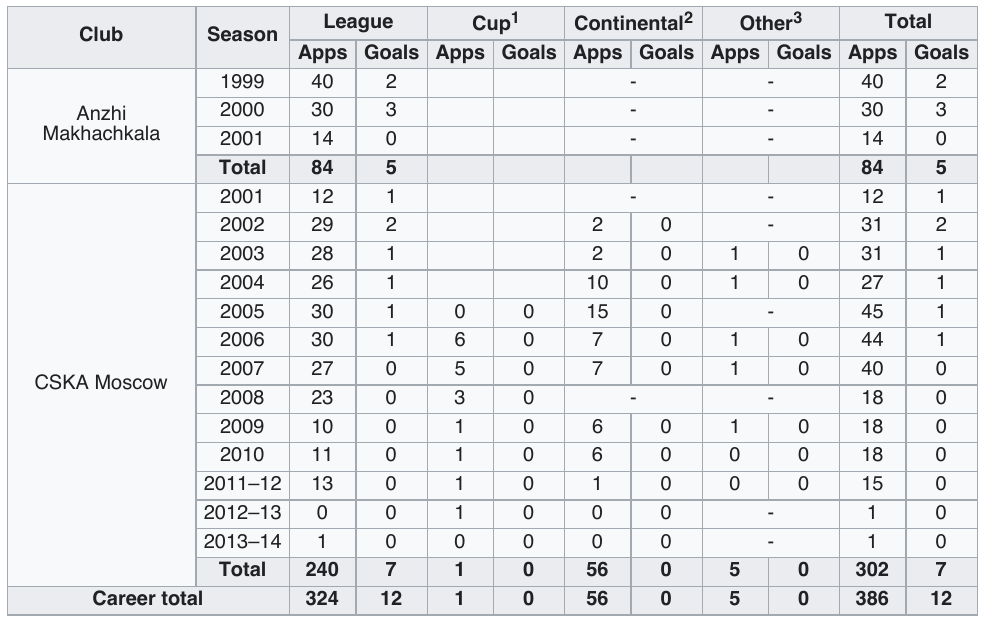}
         \caption{A sample table with a complex structure from the WikiTableQA dataset.}
         \label{fig:wiki_table_samplew_with_complex_structure}
     \end{subfigure}
     \hfill
     \begin{subfigure}[b]{\columnwidth}
         \centering
         \includegraphics[width=\columnwidth]{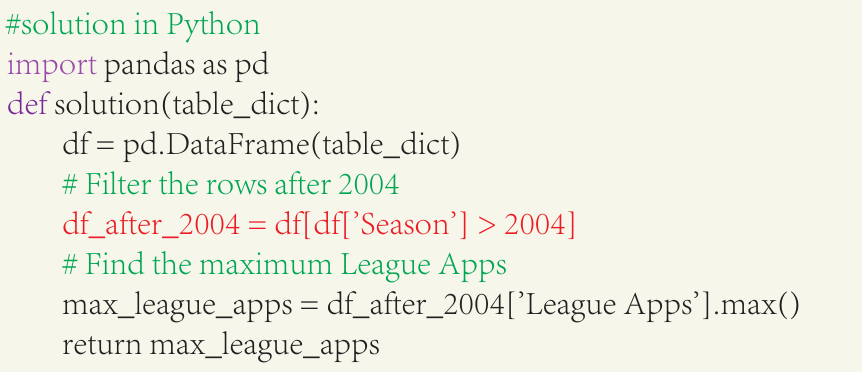}
         \caption{A sample of generated Python code to answer the question \textit{What is the maximum League Apps after 2004}.}
         \label{fig:sample_of_failure_python_code}
     \end{subfigure}
    \caption{An example of a semi-structured table and its failed Python code because of the heterogeneous data types and table's complex structure. The Python code is generated by Mixtral-8x7B. }
    \label{fig:data_sample}
\end{figure}

As mentioned in Section~\ref{sec:introduction}, many prompting methods have been proposed to enhance LLMs' capacities. For example, Chain of Thought (CoT)~\cite{wei2022chain} is a popular prompting method providing reasoning rationales to elicit LLMs' reasoning capacities. PAL~\cite{gao2023pal} and PoT~\cite{chen2023program} propose to offload the reasoning steps to Python codes. However, these typical prompting methods cannot perform well in the semi-structured Table-QA problem because of the complex structures, heterogeneous data types, and sometimes huge tables with numerous columns and rows. More specifically, prompting methods without applying programming languages must first extract the correct information from semi-structured tables before reasoning, which is challenging for LLMs, especially when the table is huge~\cite{chen2023large}. Figure~\ref{fig:method_comparision_with_benchmarks} contains a failure example of CoT, which interprets \textit{1936/37} as two years and fails to extract the correct \textit{Year} \textit{1953/54}. Similarly, programming-aided solutions, such as PAL and PoT, can also suffer from this information extraction issue if we define the relevant information as Python variables. Applying Pandas Library~\cite{reback2020pandas,mckinney-proc-scipy-2010} can alleviate this information extraction issue~\cite{multihiertt} by providing proper selection criteria, but introducing extra difficulties caused by the heterogeneous data types. Figure~\ref{fig:method_comparision_with_benchmarks} shows a PoT example using Pandas Library, which fails to run because the values in column \textit{Year} cannot be directly compared with \textit{1936}. Besides, complex structures of semi-structured tables can often lead to wrong results, especially for program-aided solutions. Figure~\ref{fig:wiki_table_samplew_with_complex_structure} shows an example from WikiTableQA~\cite{pasupat2015compositional} dataset, which contains several spanning cells across multiple columns and rows, and inconsistent data types, such as the column \textit{Season}. Even though some methods~\cite{xiao2023rethinking, fernandes2023tablestrrec, smock2022pubtables} can transform this table into a standard table by repeating table spanning cell into multiple single table cells so that SQL or Python Pandas library can process it, its structure still can lead to wrong results. For example, when an LLM is prompted to generate Python code to answer the question \textit{"What is the maximum League Apps after 2004?"}. Two \textit{Total}s in the column \textit{Season} will be compared with correct \textit{Season}s, which leads to wrong results. Besides, the generated code needs to compare the values from the column \textit{Season}, which can lead to an error of execution because $ ">" $ operation cannot be applied to the string \textit{"2011-12"} and the integer \textit{2004}, as highlighted in Figure~\ref{fig:sample_of_failure_python_code}, which is another failure example caused by the heterogeneous data types. Along with applying Python to enhance the LLMs, some studies~\cite{cheng2022binding, ye2023large} apply SQL to the Table-QA problem. However, SQL is a programming language designed for structured tables, meaning that semi-structured tables need to be transformed into structured format first so that tables can be imported into the relational databases to execute SQL queries, which is another challenging task for the semi-structured Table-QA problem discussed in this study. Besides, as pointed out by some studies~\cite{shi2020potential, cheng2022binding}, some questions are not answerable by merely using SQL. Therefore, considering the flexibility of Python and the limitations of applying SQL to the semi-structured Table-QA task, we apply Python instead of SQL in our solution.

Besides the issues of applying prompting methods to the semi-structured Table-QA problem, current studies only focus on optimizing the prediction accuracy without considering their inference cost, which is critical for our proposed GIoT system. As mentioned, some tables can be huge, which can lead to long prompts and inference time. Even though some solutions~\cite{zhang2023reactable, ye2023large} propose decomposing huge tables into sub-tables for further reasoning, they need to prompt the full huge table into the LLM first. Some studies~\cite{chen2023large} truncate the huge tables, which can drastically reduce the inference cost but introduce the risk of losing critical information in the tables and sometimes make it impossible to give the correct answer. Besides, ensemble methods, such as self-consistency decoding and majority voting, are often employed to improve the performance further~\cite{wang2022self, ye2023large, zhang2023reactable}, but also drastically increasing the inference cost simultaneously. 

Lastly, most of these studies are based on In Context Learning (ICL), using demonstrations to guide the LLM in conducting target tasks, while crafting and selecting proper demonstrations is still an open issue. Many studies~\cite{fu2022complexity, wang2023towards} pointed out that the content, number and order of demonstration can all influence the results. Therefore, we define a series of atomic operations to measure the complexity of a query question to the given table and describe the steps with the defined atomic operations, which can be a metric for the demonstration selection when crafting prompts. To mitigate the issues limiting the performance of LLMs, including the complex table structure, heterogeneous data types, huge tables, and the inherent limitations of LLMs in reasoning capacities, we propose a three-stage prompting solution containing task-planning, task-conducting and task-correction stages, as shown in Figure~\ref{fig:method_comparision_with_benchmarks}. The task-planning stage prompts the LLM to analyze the statistical information of the given table and provide programming steps, data requirements, and relevant columns to solve the query question and generate a plan. Then, the task-conducting stage first generates a default answer as the final answer when the generated Python code fails to execute and then generates the Python code based on the plan from the first stage. When a task-conducting stage fails to execute, the heterogeneous data types are usually the reason. Therefore, we also include a task-correction stage, which can generate normalization functions to normalize the data and correct the error in the task-conducting stage. It is worth mentioning that our proposed method can avoid huge tables as a part of the prompt, which can drastically reduce the number of prompting tokens when the tables are huge. 

\subsection{Proposed Prompting Method for Table-QA}
\label{sec:proposed_prompting_method_for_tqa}
\subsubsection{Overall Workflow}
\label{sec:overall_architecture}

\begin{figure*}[htp]
\begin{center}
  \includegraphics[width=\textwidth]{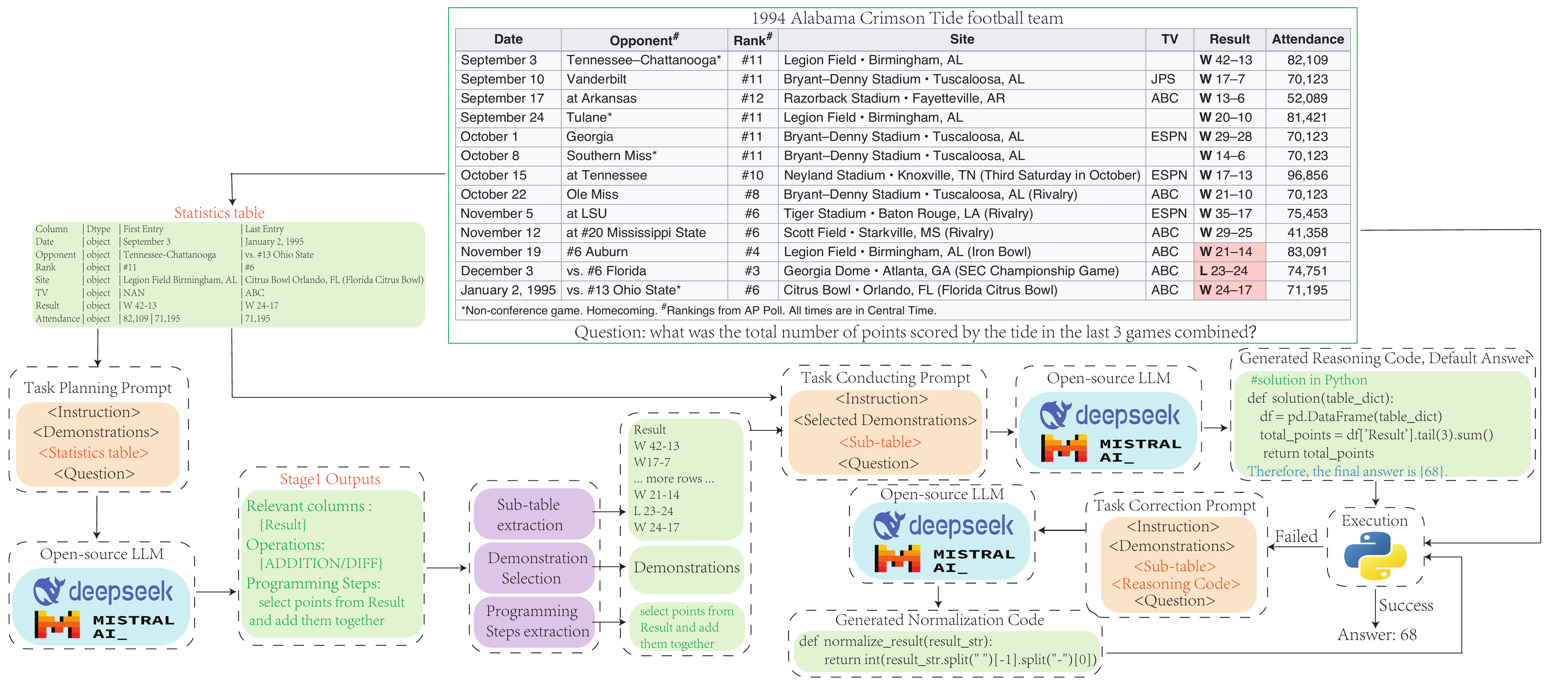}
  \caption{The workflow of the proposed prompting solution. Notably, the question and the table are from the request of an IoT device. The Python interpreter is in the Post-processing Module, and the stages of selecting demonstrations and creating prompts are in the Prompt Management Module.}
  \label{fig:overall_architecture}
\end{center}
\end{figure*}

As discussed in Sections~\ref{sec:introduction} and ~\ref{sec:problem_analysis}, we use the challenging Table-QA task as the case study of the proposed LLM-based GIoT system and propose to prompt open-source LLMs to generate Python code for the semi-structured Table-QA problem to mitigate the issues caused by complex table structures, heterogeneous data types, huge tables, and limitations of LLMs. Specifically, we propose a three-stage prompting method, including task-planning, task-conducting and task-correction stages. The proposed workflow is shown in Figure~\ref{fig:overall_architecture}, in which the question is \textit{the total number of points scored by the tide in the last 3 games combined}. To answer this question, the statistics table of the given table contains information regarding the column names, data types, and the first and last entries, which are first generated as a part of the task-planning prompt, together with instructions, demonstrations, and questions, as shown in Figure~\ref{fig:task_planning_prompt}. Then, the task-planning prompt is fed into the open-source LLM to make the reasoning plan, including Relevant Columns, Operations, and Programming Steps. It is worth mentioning that the statistics table can be far more compact than the original table when the original table is huge, which can reduce the number of prompt tokens. With the Relevant Columns, Operations and Programming Steps from the results of the task-planning step, the Relevant Columns are used to extract the contents of these columns, the Operations are used to select the demonstrations, and the Programming Steps are parts of the instructions to guide the reasoning function generation. With these processed results, the task-conducting prompt is constructed and fed into the open-source LLM to generate the reasoning Python function and a default answer, in which the default is treated as the final answer if the code fails to run. Notably, since we use the Pandas library in our solution, the original table needs to be represented as a dictionary of the list and fed into the generated function as a parameter. As running the generated Python code is almost cost-free compared with LLM inference, and some tables do not need to be normalized, we try to execute the reasoning function first. If there is no error from the execution, then the result of the execution should be the final result. However, if the execution fails, we construct the task-correction prompt containing the content of Relevant columns and reasoning code to generate the normalization function, apply the normalization to the original table, and then feed the normalized original table as the parameter to the reasoning function to run it again. For the example in Figure~\ref{fig:overall_architecture},  we need to extract the string "W 21-14", "L 23-24" and "W 24-17" first and then extract the scores "21", "23", "24" and convert them into integers, which beyond the capacities of the LLM. Therefore, the first run of the reasoning function would fail, even though its reasoning logic is correct. While the normalization function can correctly extract and convert the points from the string to integers, the second run of the reasoning function should be successful. 

\subsubsection{Demonstration Crafting and Selection}
\label{sec:demonstration_crafting_selection}
\begin{figure}[htp]
\begin{center}
  \includegraphics[width=\columnwidth]{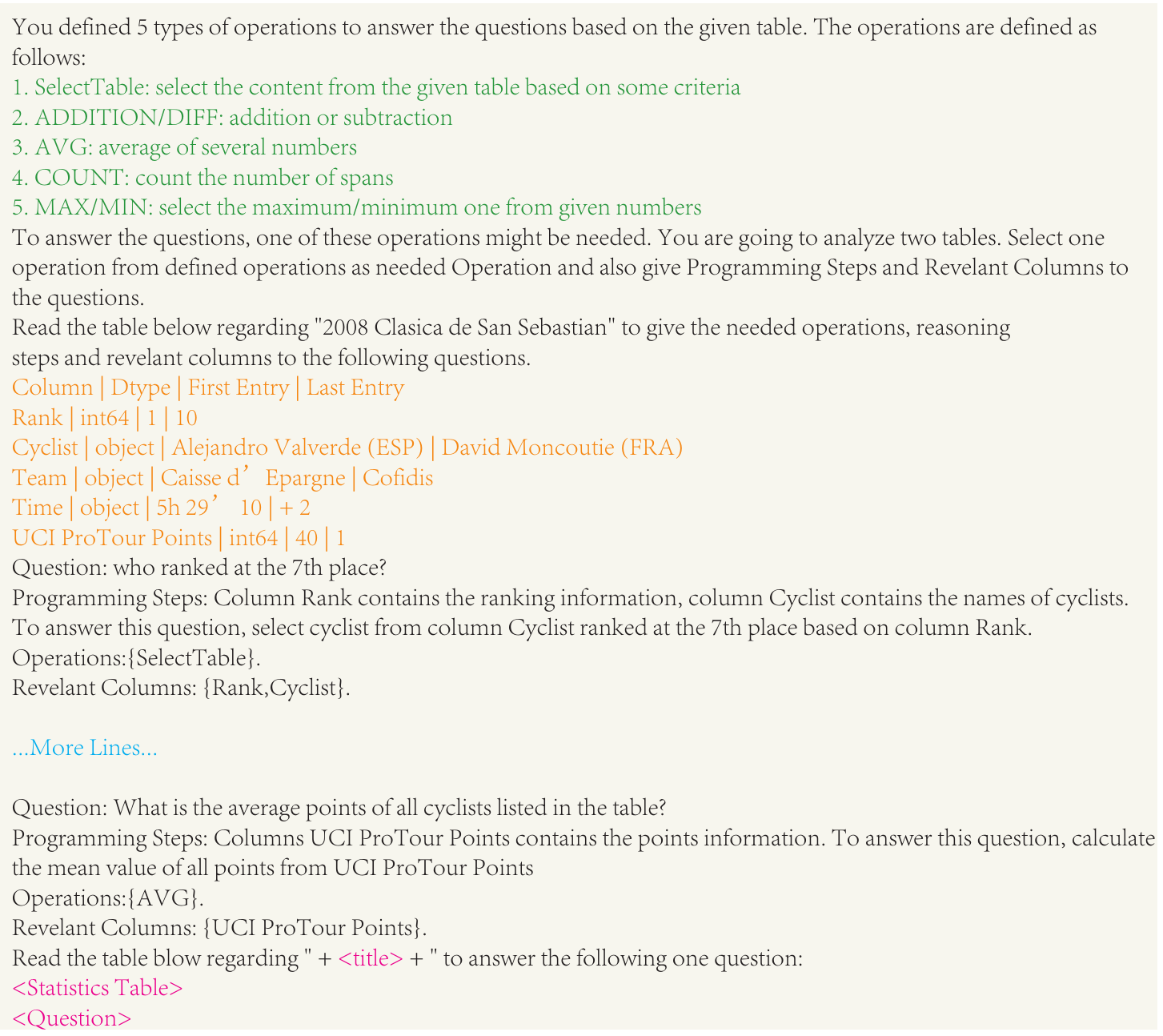}
  \caption{The task-planning prompt. The defined operations and the statistics table are highlighted with green and yellow. $<$title$>$, $<$Statistics Table$>$ and $<$Question$>$ are from the table to be analyzed.}
  \label{fig:task_planning_prompt}
\end{center}
\end{figure}

\begin{figure}[htp]
\begin{center}
  \includegraphics[width=\columnwidth]{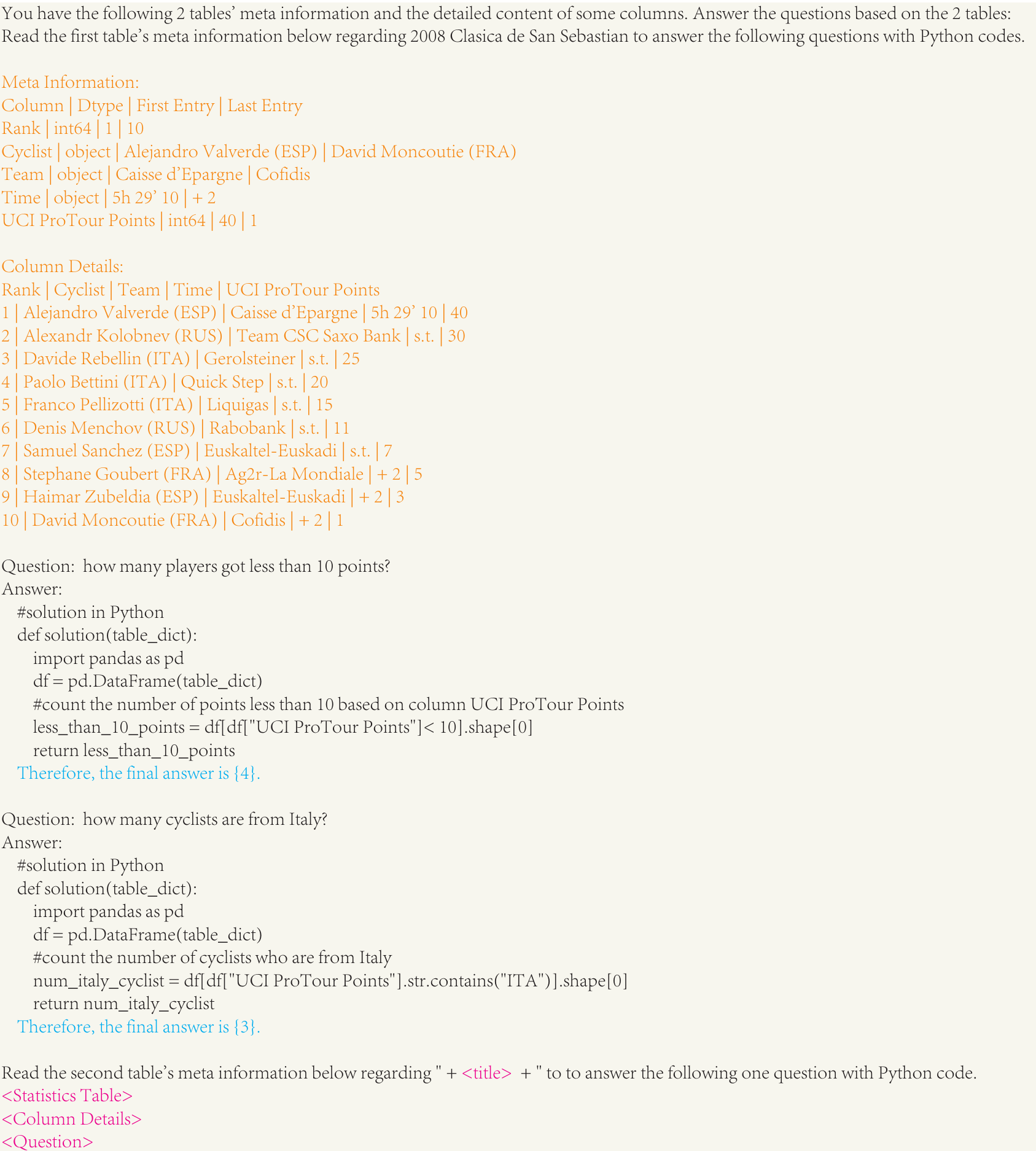}
  \caption{The task-conducting prompt. The Statistics Table and Column Details are highlighted with yellow, and the default answers are highlighted with blue. $<$title$>$, red $<$Statistics Table$>$, $<$Column Details$>$ and $<$Question$>$ are from the table to be analyzed.}
  \label{fig:task_conducting_prompt}
\end{center}
\end{figure}

As shown in Figure~\ref{fig:overall_architecture}, three prompts need to be constructed to answer a question; the demonstrations used in these three prompts are critical for the LLM's performance, especially for the task-conducting stage to generate the reasoning function. Therefore, we defined a series of atomic operations to describe the reasoning logic to answer questions, as shown in Table~\ref{tab:operations_for_demonstration_selection}. For the task-planning step, we craft a question-and-answer pair for each type of operation and instruct the LLM to generate the Relevant Columns, Operations and Programming Steps, as shown in Figure~\ref{fig:task_planning_prompt}. For the task-conducting prompts, we construct two question and reasoning function pairs for each defined operation and use operation as the metric to select these pairs. Even though we prompt the LLM to select one defined operation as output, the LLM can generate results without following the instructions. Therefore, when the Operations cannot be matched, the default setting contains all the question and function pairs. Figure~\ref{fig:task_conducting_prompt} shows an example of COUNT operation, which includes a Statistics Table, Column Details and two questions with their Python solutions and default answers. It is worth mentioning that the demonstrations and prompts discussed in this section are stored in the Task-specific Prompts Database, and using operations as metrics to select proper demonstrations is the function of the proposed Prompt Management Module, as shown in Figures~\ref{fig:giot_system} and ~\ref{fig:detailed_workflow_giot}.

\begin{table*}[!htpb]
\centering
\caption{Defined operations for Demonstration selection for code generation.}
\begin{tabular}{ c c c  } 
\hline
Operation Type & Operation Name & Description \\
\hline
\multirow{7}{*}{Reasoning} &  SelectTable & select a cell from the table based on a criteria\\
& ADDITION/DIFF & addition or subtraction   \\
& TIMES/DIVISION &   production or quotient of two numbers  \\
& AVG &   average of several numbers  \\
& COUNT &   count the number based on a criteria  \\
& MAX/MIN&  select the maximum/minimum one from given numbers   \\
\hline
\end{tabular}
\label{tab:operations_for_demonstration_selection}
\end{table*}

\section{Experiments and Analysis}
\label{sec:experiments}

\subsection{Datasets and Experimental Settings} 
\label{sec:datasets}
We evaluate our proposed prompting solution on WikiTableQA~\cite{pasupat2015compositional} and TabFact~\cite{chen2019tabfact} datasets, which are two datasets created by Wiki-tables without text context. WikiTableQA mainly contains compositional questions, such as questions requiring counting and ranking table contents. TabFact is a Fact Verification dataset, which can be treated as a special setting of a typical Table-QA problem whose answer set is $\{True, False\}$. Since the proposed solution of this study is based on ICL, which is a few-shot learning setting without any training stage, we only use the test set of these two datasets to evaluate the performance, which contains 4344 and 12828 QA pairs, respectively. TabFact dataset further categorizes the test set into simple and complex sets, which include 4219 and 8609 QA pairs, respectively. For the simple set, the QA pairs are usually obtained from a single row or record in the table, reflecting unary facts without complex logical inference. By contrast, for the complex set, the QA pairs are created by information from multiple columns and rows and derived by complex semantic operations, such as argmax, argmin, and the table records are also rewritten to include more semantic understanding. Considering the large size of TabFact dataset, some studies~\cite{ye2023large, cheng2022binding} conducted experiments one a small subset of TabFact, which contains 1,005 simple and 1,019 complex QA pairs. To compare with these studies, we also report the results on this small test subset of TabFact. 

\begin{table}[ht!]
\caption{Experimental results on WikiTableQA dataset with Exact Match Accuracy as metric.} 
\centering
\begin{tabular}{ c c c }
\hline
\label{table:results_on_wiki_tables}
LLM & Method & EM Acc\\
\hline
\multirow{2}{*}{Codex} & Binder & 61.90\\
 & Dater & 65.90 \\
\hline
\multirow{4}{*}{Mixtral-8x7B} & Direct & 53.08\\
 & CoT & 53.48\\
 & PoT & 40.40 \\
 & Tab-PoT & 63.33 \\
\hline
\multirow{4}{*}{Mistral-7B} & Direct & 27.19\\
 & CoT & 30.46\\
 & PoT & 27.66 \\
 & Tab-PoT & 52.12 \\
\hline
\multirow{4}{*}{DeepSeek-67B} & Direct & 54.72\\
 & CoT & 55.57\\
 & PoT & 43.92 \\
 & Tab-PoT & \textbf{66.78} \\
\hline
\multirow{4}{*}{DeepSeek-7B} & Direct & 33.86\\
 & CoT & 34.65\\
 & PoT & 19.61 \\
 & Tab-PoT & 40.03\\
\hline
\end{tabular}
\end{table}
As the proposed LLM-based GIoT system is designed to be deployed in a local network, we use open-source LLMs, including Mixtral-8x7B~\footnote{\url{https://huggingface.co/mistralai/Mixtral-8x7B-Instruct-v0.1}}, Mistral-7B~\footnote{\url{https://huggingface.co/mistralai/Mistral-7B-Instruct-v0.2}}, DeepSeek-67B~\footnote{\url{https://huggingface.co/deepseek-ai/deepseek-llm-67b-chat}} and DeepSeek-7B~\footnote{\url{https://huggingface.co/deepseek-ai/deepseek-llm-7b-chat}} to conduct our experiments. Specifically, we deploy the LLMs with vLLM~\cite{kwon2023efficient} on a workstation with 8 NVIDIA A100 40G GPUs. For the Mistral-7B and DeepSeek-7B models with default 16-bit parameters, one A100 40G GPU has provided enough RAM. By contrast, for the Mixtral8x7B and DeepSeek-67B with default 16-bit parameters, four A100 40G GPUs are needed for their deployment with tensor parallel.

Since the proposed solution in this study is a prompt engineering method, we include Direct Prompting, CoT~\cite{wei2022chain}, PoT~\cite{chen2023program}, Binder~\cite{cheng2022binding} and Dater~\cite{ye2023large} as benchmarks, in which Binder and Dater are two solutions leveraging SQL. For the implementation of benchmark methods, we use the implementation of TableCoT\footnote{\url{https://github.com/wenhuchen/TableCoT}}~\cite{chen2023large} for the Direct Prompting and CoT. We re-implemented the PoT method following the example prompts reported in PoT~\cite{chen2023program}. The results of Dater and Binder are directly from study~\cite{ye2023large}. We use a simple beam search decoding for the experiments. It is worth mentioning that self-consistency and more complex sampling decoding methods are often used in state-of-the-art techniques, such as Dater and Binder, to improve the quality of generated texts, which can introduce significant inference overhead. At last, even though we employ ICL to provide demonstrations to guide the LLM output of the final results in a "\{\}," sometimes LLMs can fail to follow this output format. Therefore, we use a simple answer alignment step to post-process the results with incorrect formats. More specifically, we employ a direct prompting method by providing a few demonstrations containing the question, answer and formatted answer following TableCoT~\cite{chen2023large}. It is worth mentioning that we use the default precision of parameters, namely bfloat16, for the LLMs in this section. 

We term our proposed prompting solution as Tab-PoT, and the experimental results are shown in Table~\ref{table:results_on_wiki_tables} and Table~\ref{table:results_on_tabfact}. The experimental results show that our proposed prompting solution can perform competitively compared with state-of-the-art methods. The Tab-PoT with DeepSeek-67B can achieve state-of-the-art performance, and the Tab-PoT with both Mixtral-8x7B and DeepSeek-67B can improve the original PoT method by at least 22\% on the WikiTableQA dataset. Besides, applying LLMs with a larger number of parameters can significantly improve performance. The CoT does not show many benefits in improving performance compared with direct prompting on the WikiTableQA dataset while consistently improving the performance on the TabFact dataset.

\begin{table*}[ht!]
\caption{Experimental results on TabFact dataset with Exact Match Accuracy as metric. $full$ and $small$ mean the full and small versions of TabFact dataset.} 
\centering
\begin{tabular}{ c c | c c c | c c c }
\hline
\label{table:results_on_tabfact}
LLM & Method & Simple$_{full}$ & Complex$_{full}$ & All$_{full}$ & Simple$_{small}$ & Complex$_{small}$ & All$_{small}$\\
\hline
\multirow{2}{*}{Codex} & Binder & - & - & - & - & - & 85.10 \\
 & Dater & - & - & - & 91.20 & 80.00 & 85.60 \\
\hline
\multirow{4}{*}{Mixtral-8x7B} & Direct & 80.59 & 69.98 & 73.47 & 81.29 & 70.56 & 75.89\\
 & CoT & 83.53 & 74.94 & 77.77 & 86.07 & 74.19 & 80.09 \\
 & PoT & 73.33 & 69.07 & 70.47 & 76.02 & 70.66 & 73.32 \\
 & Tab-PoT & 86.49 & 76.06 & 79.49 & 88.36 & 75.07 & 81.67 \\
\hline
\multirow{4}{*}{Mixtral-7B} & Direct & 73.57 & 64.83 & 67.70 & 73.13 & 62.71 & 67.89\\
 & CoT & 73.31 & 67.52 & 69.43 & 73.73 & 67.12 & 70.41\\
 & PoT & 66.11 & 63.58 & 64.41 & 65.97 & 66.54 & 66.25\\
 & Tab-PoT & 77.48 & 66.83 & 69.75 & 76.82 & 66.93 & 71.84\\
\hline
\multirow{4}{*}{DeepSeek-67B} & Direct & 84.36 & 74.19 & 77.53 & 84.68 & 72.62 & 78.61 \\
 & CoT & 87.44 & 78.00 & 81.10 & 88.46 & 76.84 & 82.61\\
 & PoT & 74.43 & 71.79 & 72.65 & 76.32 & 72.72 & 74.51\\
 & Tab-PoT & \textbf{90.09} & \textbf{78.91} & \textbf{82.58} & \textbf{91.34} & \textbf{80.27} & \textbf{85.77}\\
\hline
\multirow{4}{*}{DeepSeek-7B} & Direct & 59.16 & 55.42 & 56.65 & 59.50 & 55.94 & 57.71\\
 & CoT & 69.31 & 61.18 & 63.85 & 70.65 & 59.76 & 65.17\\
 & PoT & 63.71 & 58.26 & 60.06 & 64.88 & 58.98 & 61.91\\
 & Tab-PoT & 70.42 & 62.13 & 64.86 & 70.75 & 61.04 & 65.86\\
\hline
\end{tabular}
\end{table*}

\subsection{Discussion and Analysis}
\label{sec:discussion_analysis}

\subsubsection{Ablation Study}
\label{abliation_study}
As discussed in Section~\ref{sec:proposed_method}, our proposed prompting solution consists of three stages: task-planning, task-conducting and task-correction. The task-planning stage can output the relevant columns and reasoning steps to answer the query question, which is also a step of table decomposition and question decomposition that can be applied to the conventional PoT. In the task-conducting stage, we prompt the LLM to generate the Python code and a default answer. The default answer is the final answer when the Python code fails to run even after the task-correction stage. Since the default answer is generated after the Python code, it can be treated as an implicit CoT where the Python code is the reasoning rationales in the CoT. Finally, the third stage generates normalization functions to correct the errors in the Python code caused by the heterogeneous data types, which rely on the relevant columns generated by the first stage. Therefore, in this section, we conduct four ablation experiments by applying task-planning, default answer in task-conducting, task-planning and task-correction, and task-planning and default answer in task-conducting to the conventional PoT. The experimental results are shown in Table~\ref{table:ablation_result}, where Plan, Correction and Default represent applying task-planning, task-correction and the default answer in task-conducting stages. We use Mixtral-8x7B as the LLM, and the experimental results demonstrate that each of the proposed three components can significantly improve the PoT baseline.

\begin{table}[ht!]
\caption{Ablation study results on the WikiTableQA dataset with Exact Match Accuracy as metric.} 
\centering
\begin{tabular}{ c c c c | c}
\hline
\label{table:ablation_result}
Model & Plan & Correction & Default & EM Acc\\
\hline
PoT &   &  &  & 40.40\\
Ablation 1& \checkmark &  & & 46.52\\
Ablation 2&  &  & \checkmark & 53.66 \\
Ablation 3& \checkmark & \checkmark &  & 53.31\\
Ablation 4& \checkmark &  & \checkmark & 58.43\\
Tab-PoT & \checkmark & \checkmark & \checkmark & 63.33\\
\hline
\end{tabular}
\end{table}

\subsubsection{The impact of quantization}
\label{sec:quantization_analysis}
Since the LLMs usually have very high hardware requirements for inference, quantization methods are widely used to compact LLMs using lower precision parameters. In this section, we conduct experiments to compare the performance of quantization versions of LLMs. Specifically, similar to the previous section, we also use Mixtral-8x7B to conduct experiments on the WikiTableQA dataset and compare its 16-bit, 8-bit and 4-bit versions of applying the proposed Tab-PoT solution and the experimental results are shown in Table~\ref{table:quantization_comparison}. For our proposed Tab-Pot, even though applying quantization methods can lead to worse performance, the performance of 8-bit and 4-bit versions is still competitive. It is worth mentioning that the 4-bit version shares a similar RAM footprint with the Mistral-7B model but achieves much higher performance, as shown in Table~\ref{table:quantization_comparison} and Table~\ref{table:results_on_wiki_tables}. 

\begin{table}[ht!]
\caption{The impact of LLM quantization methods. Notably, the LLM used in this table is Mixtral-8x7B.} 
\centering
\begin{tabular}{ c c c c c c}
\hline
\label{table:quantization_comparison}
 \multirow{2}{*}{\#GPU} & Model & Parameter &   EM  & AVG Prompt & AVG Completion\\
 &  Size & Precision& Acc & Throughput & Throughput\\

\hline
4 & 87G & 16-bit & 63.33 & 743.8 tokens/s & 58.8 tokens/s\\
4 & 48G & 8-bit & 62.20 & 553.7 tokens/s & 39.7 tokens/s\\
2 & 48G & 8-bit & 62.20 & 377.6 tokens/s & 26.5 tokens/s\\
4 & 25G& 4-bit & 60.52 & 581.3 tokens/s & 49.3 tokens/s\\
2 & 25G& 4-bit & 60.52 & 427.7 tokens/s & 35.9 tokens/s\\
1 & 25G& 4-bit & 60.52 & 286.8 tokens/s & 24.3 tokens/s\\    
\hline
\end{tabular}
\end{table}

\subsubsection{Analysis on different implementations of PoT}
\label{sec:pot_implementation_analysis}
\begin{figure}[htp]
\begin{center}
  \includegraphics[width=\columnwidth]{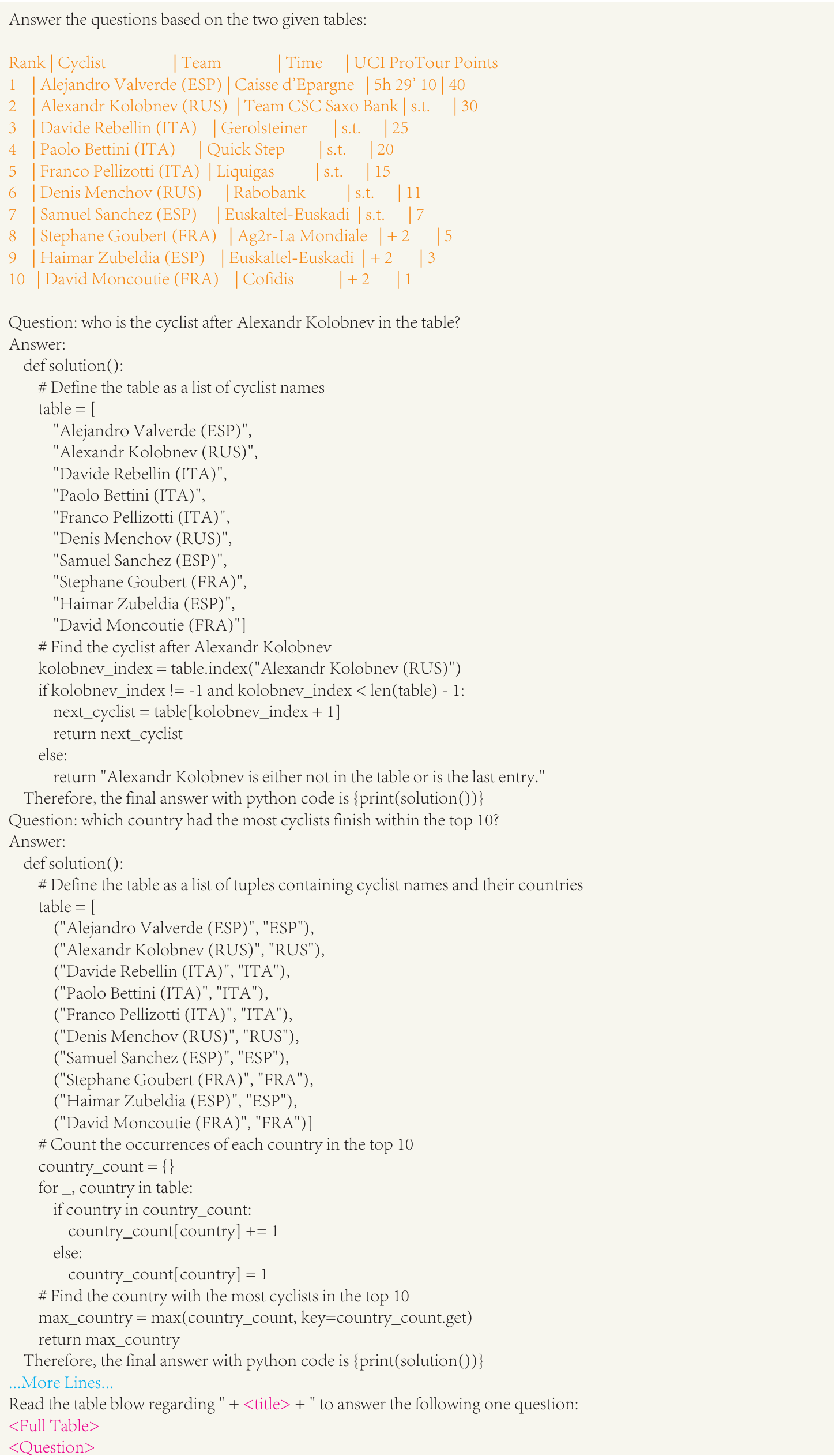}
  \caption{PoT implementation of applying Python Standard Library. Some lines are omitted due to the limited page.}
  \label{fig:pot_with_std_lib_prompting}
\end{center}
\end{figure}

\begin{figure}[htp]
\begin{center}
  \includegraphics[width=\columnwidth]{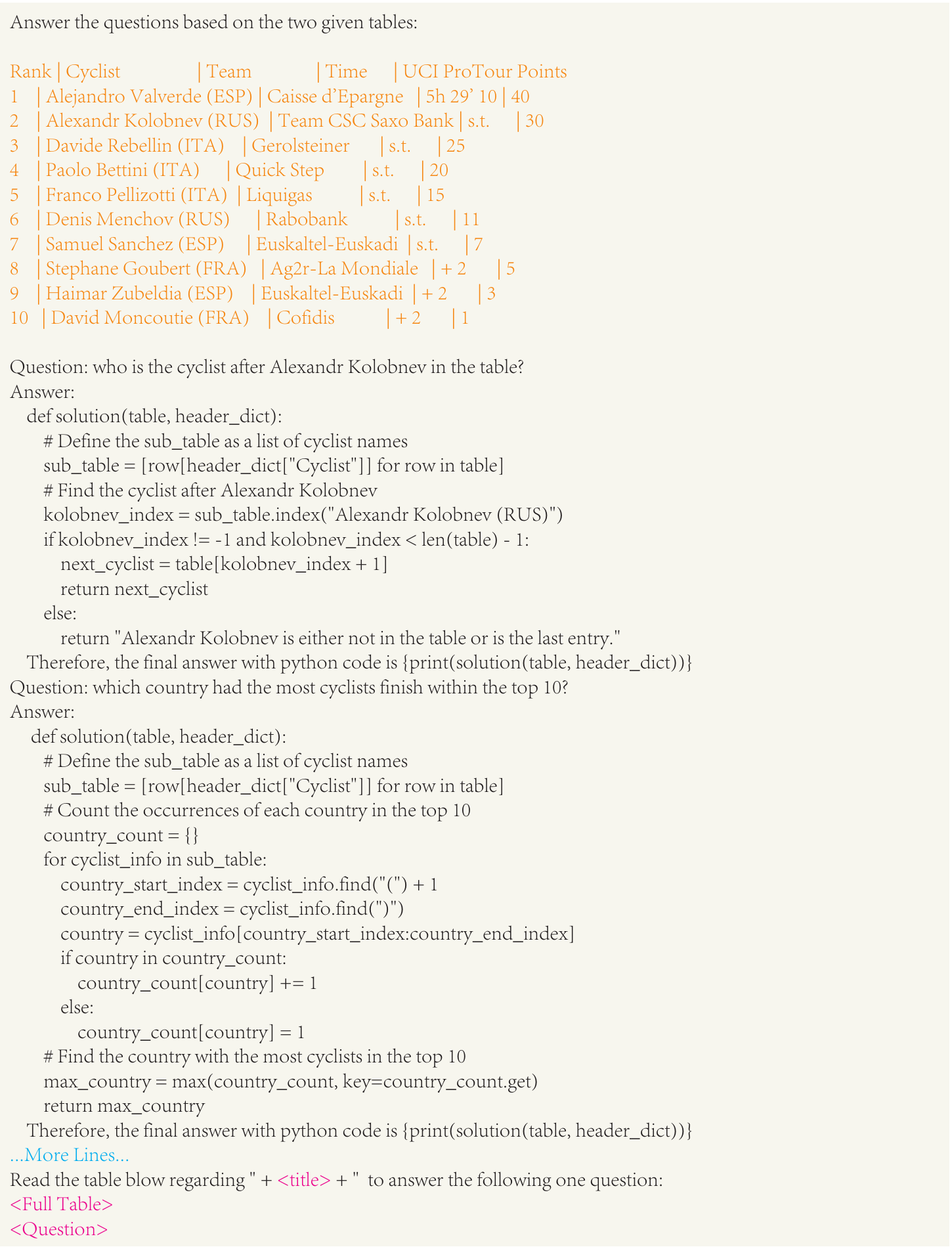}
  \caption{PoT implementation of applying Python Standard Library with parameters. Some lines are omitted due to the limited page.}
  \label{fig:pot_with_std_lib_prompting_parameters}
\end{center}
\end{figure}

\begin{figure}[htp]
\begin{center}
  \includegraphics[width=\columnwidth]{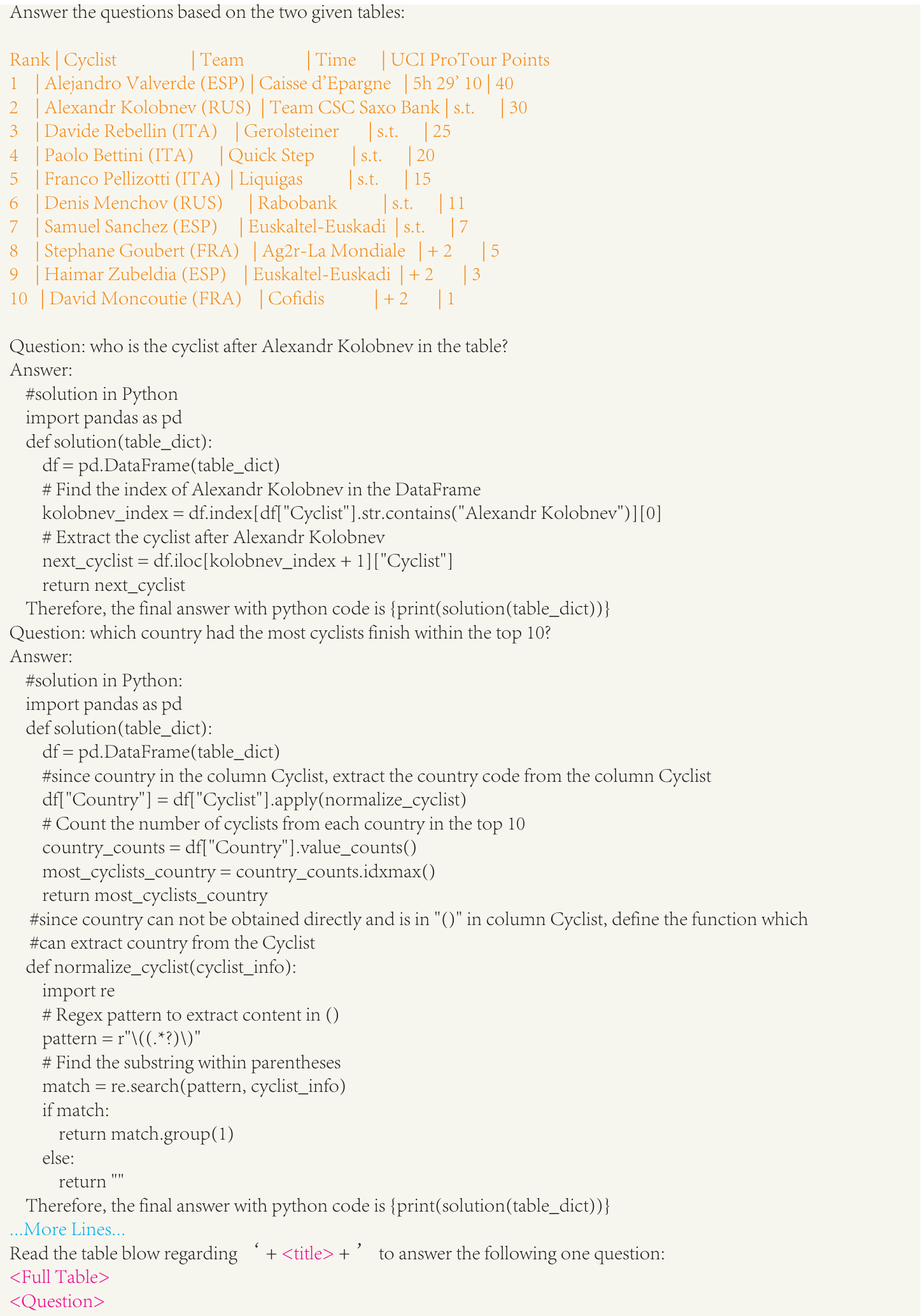}
  \caption{PoT implementation of applying Python Pandas Library. Some lines are omitted due to the limited page.}
  \label{fig:pot_with_pandas_prompting}
\end{center}
\end{figure}

Since Python is a flexible programming language that can implement a function with multiple implementations. In this section, we discuss the differences among these different types of implementations. More specifically, one straightforward implementation is using the Python Standard Library and following the extraction and reasoning steps, as shown in Figure~\ref{fig:pot_with_std_lib_prompting}. This method selects relevant data from the table, defines the data with a List of Tuples, and then conducts reasoning over the defined List of Tuples. One obvious drawback of this implementation method is that it needs to repeat the relevant columns in the Python code, which can be very large when the table contains a large number of rows, leading to more inference time. Therefore, a refined method can use the table as the parameter of the solution function, then as shown in Figure~\ref{fig:pot_with_std_lib_prompting_parameters}. At last, since Pandas is a widely used Python library to process tabular data, we can also use Pandas to finish the reasoning tasks with a table dictionary as the input, as shown in Figure~\ref{fig:pot_with_pandas_prompting}. We conduct experiments on the WikiTableQA dataset to compare the performance of these three types of implementations. Even though the implementation of applying Python Standard Library can show some benefits regarding the EM Accuracy, it requires more Prompt Tokens and Completion Tokens, as shown in Table~\ref{tab:comparisions_of_pot_implementations}, because this implementation needs to extract relevant from the table directly and define them as a Python dictionary, List or variables. On the other hand, both solutions introducing function parameters can reduce the number of prompting tokens and completion tokens, and applying Pandas Library can achieve better performance than using the standard library.

\begin{table}[ht!]
\caption{Comparisons of three types of three different types of implementations for the PoT method.} 
\centering
\begin{tabular}{ c c c c}
\hline
\label{tab:comparisions_of_pot_implementations}
\multirow{2}{*}{Method} & \multirow{2}{*}{EM Acc}& \#AVG Prompt & \#AVG Completion\\
 & & Tokens & Tokens\\
\hline
STDLib & 44.96 & 2365 & 732\\
STDLib-Para & 31.17 & 2157 & 114\\
Pandas & 40.40 & 2241 & 88\\
\hline
\end{tabular}
\end{table}

\subsubsection{Analysis on inference cost}
\begin{figure}[htp]
\begin{center}
  \includegraphics[width=\columnwidth]{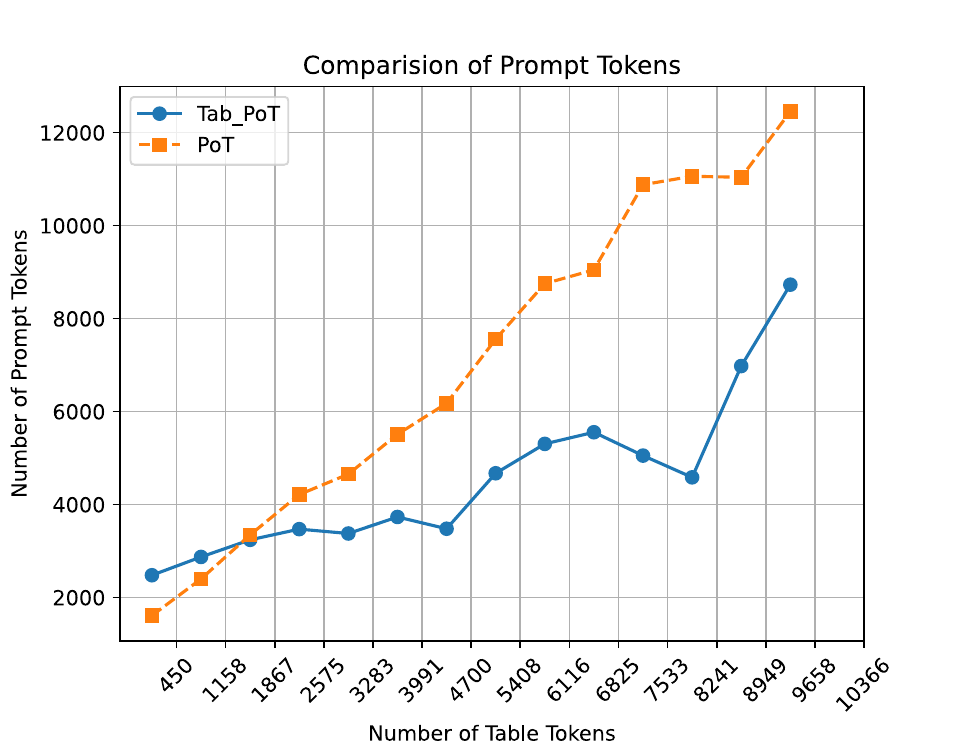}
  \caption{Comparison of Prompting Tokens between PoT and Tab\_PoT.}
  \label{fig:comparsion_of_prompting_tokens}
\end{center}
\end{figure}

Since the inference cost is highly correlated with the number of tokens, we use the prompt tokens and generated tokens as the metrics to measure the inference cost in this section. Therefore, we calculate the Average Prompt Tokens and Completion Tokens on the WikiTableQA dataset. As shown in Table~\ref{tab:token_cost}, the proposed Tab-PoT can introduce some overhead compared with other methods regarding the average prompt tokens and average completion tokens on the WikiTableQA dataset. Since the proposed Tab-Pot contains three stages at most, each including instructions and demonstrations, it can reduce the number of Prompt Tokens only when the input table is huge. We group the number of table tokens into 15 bins and plot the relation between the number of table tokens and the prompt tokens on the WikiTableQA dataset. When the number of a table's tokens is larger than around 1867, our proposed Tab\_PoT can use fewer prompt tokens than PoT, which means fewer computation operations and less inference time, as shown in Figure~\ref{fig:comparsion_of_prompting_tokens}. Since the WikitTableQA dataset contains a large portion of tables whose number of tokens is smaller than 1158, the average prompt tokens of the proposed Tab-PoT is still larger than the one of PoT overall, as shown in Table~\ref{tab:token_cost}. As pointed out by some studies~\cite{chen2023large}, the LLM can perform well on small tables, meaning that we can easily extend the proposed Tab-PoT with other methods, such as CoT, by applying a threshold regarding the size of the input table, to reduce the number of prompt tokens and maintain the performance simultaneously. It is worth mentioning that the price of prompt tokens and completion tokens are different when using commercial LLMs, such as GPT-4~\cite{OpenAI2023GPT4TR}.

\begin{table}[ht!]
\caption{Comparisons of Average Prompt Tokens and Completion Tokens.} 
\centering
\begin{tabular}{ c c c}
\hline
\label{tab:token_cost}
Method & \#AVG Prompt Tokens & \#AVG Completion Tokens\\
\hline
 Direct & 1405 & 10 \\
 CoT  & 1599 & 44 \\
 PoT  & 2241 & 88\\
 Tab-PoT & 2685 & 192\\
\hline
\end{tabular}
\end{table}

As mentioned in Section~\ref{sec:datasets}, we deployed the LLMs with vLLM~\cite{kwon2023efficient} on a workstation with 8 NVIDIA A100 40G GPUs. We utilized a single GPU for the Mistral-7B and DeepSeek-7B and four GPUs for the Mixtral-8x7B and DeepSeek-67B. The throughput performances of each model (with respect to prompt and completion) are summarized in Table~\ref{tab:inference_speed}. As shown in Table~\ref{tab:inference_speed}, the throughput performance of the DeepSeek-67B is significantly lower than other models for both prompt and completion, which is caused by the overhead of using tensor parallelism. Specifically, we observed that the number of components of the DeepSeek-67B deployed in the CPUs is much larger than that of Mixtral-8x7B for the tensor parallelism, making it much slower than Mixtral-8x7B which also needs to be deployed using tensor parallelism. By contrast, the Mistral-7B and DeepSeek-7B, both of which can be deployed in a single GPU, can show similar throughput performances.
At last, it is worth mentioning that the inference overheads caused by the Prompt Management Module and the Post-processing Module are negligible, because they can be finished in milliseconds for our cases, while the LLM inference usually needs at least a few seconds.

\begin{table}[ht!]
\caption{Comparisons of inference speeds of LLMs. } 
\centering
\begin{tabular}{ c c c c}
\hline
\label{tab:inference_speed}
\multirow{2}{*}{Model} & \multirow{2}{*}{\#GPU} & AVG Prompt  & AVG Completion \\
  & & Throughput & Throughput\\
\hline
Mixtral-8x7B & 4 & 743.8 tokens/s & 58.8 tokens/s\\
Mistral-7B & 1 & 633.0 tokens/s & 57.7 tokens/s \\
DeepSeek-67B & 4 & 320.7 tokens/s & 20.1 tokens/s\\
DeepSeek-7B & 1 & 705.2 tokens/s & 59.4 tokens/s \\
\hline
\end{tabular}
\end{table}

\section{Conclusion and Future Work}
\label{sec:conclusion}
In this study, we propose an LLM-based GIoT system, which can be deployed in a local network setting to address the security concerns of many scenarios. To demonstrate the proposed LLM-based GIoT system, we use a challenging semi-structured Table-QA problem as a case study and propose a three-stage prompting solution to alleviate the issues caused by complex structures, heterogeneous data types, huge tables, and LLMs' limitations. The proposed prompting solution uses a statistics table in the first stage and sub-tables in the following stages, which can reduce the inference cost and improve the performance when the original table is huge. We define a series of atomic operations to guide the demonstration crafting and selection, which can reduce reasoning errors. Besides, we use the task-correction stage to correct the failure code caused by the heterogeneous data types and use a default answer as the final answer when the generated Python code fails to run even after the task-correction step, which can be caused by the complex structure or the limitations of the LLM. As demonstrated in Section~\ref{sec:experiments}, designing tailored prompting methods can improve the performance of open-source LLMs, achieving state-of-the-art performance, and the proposed LLM-based GIoT system can be easily extended by adding task-specific prompt instructions and demonstrations to the system. Besides, as the proposed GIoT system is designed to deploy in an edge server, applying quantization to the LLM can be a good option to reduce the hardware requirements, as discussed in Section~\ref{sec:quantization_analysis}. In this study, we focus on text data, while IoT devices can generate data in multiple data types, such as time series and images. Therefore, extending the current system to handle data in various modalities can be a further direction. Besides extending the system to multiple modalities, improving computational efficiency and reducing hardware requirements are also critical to the system. Therefore, the caching mechanisms which caches intermediate results or frequently used outputs, model pruning which can reduce the size of LLM models, and many others are also promising directions.


%





\ifCLASSOPTIONcaptionsoff
  \newpage
\fi



%



%

\section*{Acknowledgement}
This work is supported in part by the Natural Sciences and Engineering Research Council of Canada (NSERC) under the CREATE TRAVERSAL Program and NSERC DISCOVERY program, and also in part by the National Research Foundation, Singapore, and Infocomm Media Development Authority under its Future Communications Research \& Development Programme, AI Singapore Programme (FCP-NTU-RG-2022-010 and FCP-ASTAR-TG-2022-003), Singapore Ministry of Education (MOE) Tier 1 (RG87/22), and the NTU Centre for Computational Technologies in Finance (NTU-CCTF).
\bibliographystyle{IEEEtran}

\end{document}